\documentclass[twocolumn]{article} 

\usepackage[margin=1in]{geometry}

\usepackage{times}

\usepackage{graphicx}
\usepackage{amsmath}
\usepackage{booktabs}
\usepackage{subfigure}
\usepackage{hyperref}
\usepackage{enumitem}
\usepackage{amsfonts}
\usepackage{amssymb}
\usepackage{amsthm}
\newtheorem{definition}{Definition}[section]

\usepackage{multirow}
\usepackage{dsfont}
\usepackage[ruled]{algorithm2e}
\usepackage[table,xcdraw]{xcolor}
\usepackage{natbib}

\usepackage{todonotes}

\begin{document}
\title{Bayesian Modeling of Intersectional Fairness:\\ The Variance of Bias}
\author{James Foulds, Rashidul Islam, Kamrun Keya, Shimei Pan\\
	University of Maryland, Baltimore County
}
\date{}
\maketitle

\begin{abstract}
  Intersectionality is a framework that analyzes how interlocking systems of power and oppression affect individuals along overlapping dimensions including race, gender, sexual orientation, class, and disability.  Intersectionality theory therefore implies it is important that fairness in artificial intelligence systems be protected with regard to multi-dimensional protected attributes.  However, the measurement of fairness becomes statistically challenging in the multi-dimensional setting due to data sparsity, which increases rapidly in the number of dimensions, and in the values per dimension.  We present a Bayesian probabilistic modeling approach for the reliable, data-efficient estimation of fairness with multi-dimensional protected attributes, which we apply to two existing intersectional fairness metrics.  Experimental results on census data and the COMPAS criminal justice recidivism dataset demonstrate the utility of our methodology, and show that Bayesian methods are valuable for the modeling and measurement of fairness in an intersectional context.
\end{abstract}

\section{Introduction}
\label{intro}

With the rising influence of machine learning algorithms on many important aspects of our daily lives, there are growing concerns that biases inherent in data can lead the behavior of these algorithms to discriminate against certain populations \citep{angwin2016machine,barocas2016big,berk2017fairness,bolukbasi2016man,dwork2012fairness,executive2016big,noble2018algorithms}. 
In recent years, substantial research effort has been devoted to the development and enforcement of mathematical definitions of bias and fairness in machine learning algorithms \citep{dwork2012fairness,hardt2016equality,kusner2017counterfactual,kearns2018preventing}.

\begin{table}[t]
	\centering
	\small
	\begin{tabular}{llll}
		\toprule
		Protected attributes & gender & gender, & gender,\\
		&        & nationality & nationality,\\
		&	      &             & race \\
		Median $\#$ instances& 14,719 & 5,195     & 172 \\                     
		Minimum $\#$ instances   & 9,216 & 963 & 5\\
		\bottomrule	
	\end{tabular}
	\caption{Number of instances at each intersection of the protected attributes' values, UCI Adult census dataset.  
	\label{tab:adultIntersectionalCounts}}
\end{table}

In this work, our guiding principle for fairness is \emph{intersectionality}, the core theoretical framework underlying the third-wave feminist movement \citep{crenshaw1989demarginalizing,collins2002black}.
Intersectionality theory states that racism, sexism, and other social systems which harm marginalized groups have interlocking effects, such that the lived experience of, e.g., Black women, is very different than that of, e.g., white women.  We therefore focus on fairness scenarios where there are \emph{multiple protected attributes}, such as gender, race, and sexual orientation.

While fairness methods have been extended to multiple protected attributes \citep{kearns2018preventing,hebert-johnson2018multicalibration, foulds2018intersectional}, data sparsity rapidly becomes an issue as the number of dimensions (and their number of distinct values) increases, leading to uncertainty in the measurement of fairness.
For example, Table \ref{tab:adultIntersectionalCounts} shows how the number of instances per value at the intersections of the protected attributes, and especially the minimum of these counts, decreases as more protected attributes are introduced, on the UCI Adult census dataset \citep{kohavi-nbtree}. It may be difficult, for instance, to estimate the overall behavior of a classifier on those individuals who are indigenous women from foreign countries, due to a lack of recorded data on such individuals.  To detect intersectional discrimination, we need to measure the system's behavior on potentially small intersectional groups, which is unreliable to estimate in the resulting ``small $N$'' regime  \citep{roth2006modeling}.

The goal of this work, therefore, is to address the challenge of \emph{reliably modeling and measuring fairness in an intersectional context, despite data sparsity.}  While small data uncertainty \citep{roth2006modeling}, intersectionality \citep{buolamwini2018gender, foulds2018intersectional}, and multiple attribute definitions \citep{kearns2018preventing,hebert-johnson2018multicalibration,foulds2018intersectional} have been studied, we are first to consider them concurrently.

The majority of the research on fairness in AI to date has focused on the development of learning algorithms which enforce fairness metrics \citep{dwork2012fairness,zemel2013learning,hardt2016equality, bolukbasi2016man,kusner2017counterfactual,berk2017convex}.  Here, we instead focus on accurately \emph{measuring} the unfairness of a system or dataset.  Fairness measurement is crucial when engineering AI systems for deployment \citep{speicher2018unified}.  It is essential for determining whether disparities in system behavior meet legal thresholds for discrimination \citep{roth2006modeling}.  And it is integral to investigative reporting on disparate behavior of existing AI systems, which promotes awareness and can ultimately lead to the rectification of algorithm injustice \citep{angwin2016machine,buolamwini2018gender, raji2019actionable}.
Our primary contributions are:
\begin{enumerate}
	\item We propose a Bayesian probabilistic modeling framework for reliably estimating fairness and its uncertainty in the data-sparse intersectional regime.
	\item We instantiate the proposed framework with four statistical models, each with a different bias and variance tradeoff, including a novel hierarchical extension of Bayesian logistic regression which is potentially an appropriate choice for this setting.  We further propose a Bayesian model averaging approach which leverages all of the models together.
	\item We study the behavior of our Bayesian models on criminal justice, census, and synthetic data.  Our results demonstrate the importance of the Bayesian modeling approach in an intersectional context.
\end{enumerate}

The remainder of the paper is structured as follows.  We begin by discussing intersectionality theory, which motivates our multi-dimensional approach to fairness, and describe two intersectional fairness metrics from the literature \citep{foulds2018intersectional, kearns2018preventing}.  
Next, we propose Bayesian probabilistic models for estimating these (and other) fairness metrics in the multi-dimensional fairness regime.
We then empirically study the behavior of the models in estimating the intersectional fairness metrics, and showcase their real-world application with a case study on the COMPAS recidivism dataset.
Finally, we conclude with a discussion of the practical implications of our work.

\section{Background and Motivation: Intersectionality and AI Fairness}
\label{sec:intersectionality}
\emph{Intersectionality} is a critical lens for analyzing how unfair processes in society, such as sexism and systemic racism, affect certain groups.  The term was original introduced by  \cite{crenshaw1989demarginalizing}, who studied how the combined harms of such \emph{systems of oppression} affect Black women, who are simultaneously affected by sexism, racism, and other related disadvantages \citep{truth1851aint,collective1977black}.  In its more general form, advanced by \cite{collins2002black} and others, intersectionality theory posits that \emph{individuals at the intersection of multiple protected categories}, along lines of gender, race, social class, disability, and so on, are harmed by overlapping systems of oppression.

In an AI fairness context, this implies that fairness should be enforced at the intersections of multiple protected attributes \citep{buolamwini2018gender,foulds2018intersectional}.  Here, we consider several existing fairness definitions which are appropriate in an intersectional context.  
\subsection{Differential Fairness}
Differential fairness \citep{foulds2018intersectional} is a definition specifically motivated by intersectionality, which aims to ensure equitable treatment by an algorithm for \emph{all intersecting subgroups} of a set of protected categories.   We use the notation of \cite{kifer2014pufferfish} for all definitions we consider.  Let $M(x)$ be an algorithmic mechanism which takes an individual's data $x$ and assigns them an outcome $y$, e.g. whether or not the individual was awarded a loan.  Let  $S_1, \ldots, S_p$ be discrete-valued protected attributes, $A = S_1 \times S_2 \times \ldots \times S_p$, and let $\theta$ be the distribution which generates $x$.
\begin{definition}  (Differential Fairness)
	A mechanism $M(x)$ is $\epsilon$-\emph{differentially fair (DF)} with respect to $(A, \Theta)$ if for all $\theta \in \Theta$ with $x \sim \theta$, and $y \in \mbox{Range}(M)$,
	\begin{equation}
	e^{-\epsilon} \leq \frac{P_{M, \theta}(M(x) = y|s_i, \theta)}{P_{M, \theta}(M(x) = y|s_j, \theta)}\leq e^\epsilon \mbox{ ,} \label{eqn:DF}
	\end{equation}
	for all   $(s_i, s_j) \in A \times A$ where $P(s_i|\theta) > 0$, $P(s_j|\theta) > 0$.
	\label{def:DF}
\end{definition}
In Definition \ref{def:DF}, $s_i$, $s_j \in A$ are tuples of \emph{all} protected attribute values, e.g. gender, race, and nationality.  If all of the $P_{M, \theta}(M(x) = y|s, \theta)$ probabilities are equal for each group $s$, across all outcomes $y$ and distributions $\theta$, $\epsilon = 0$, otherwise $\epsilon > 0$.  \cite{foulds2018intersectional} proved that this definition guarantees fairness protections for all subsets of the protected attributes, e.g. if all intersections of gender and race are protected (e.g. Black women), then gender (e.g. women) and race (e.g. white people) are separately protected, a property which is consistent with the ethical principles of intersectionality theory.  \cite{foulds2018intersectional} further proposed a variant definition which only considers the increase in unfairness by the algorithm, over the unfairness in the original data.
\begin{definition} (DF Bias Amplification)
	A mechanism $M(x)$ satisfies $(\epsilon_2 - \epsilon_1)$-\emph{DF bias amplification} with respect to $(A, \Theta, D, \mathcal{M})$ if it is $\epsilon_2$-DF and $D$ is a labeled dataset which is $\epsilon_1$-DF w.r.t. a model $\mathcal{M}$ which was trained on $D$ to estimate $P(y|s)$ in the data.
	\label{def:biasAmplification}
\end{definition}

\subsection{Subgroup Fairness}
\cite{kearns2018preventing} proposed multi-attribute fairness definitions which aim to prevent \emph{fairness gerrymandering} at the intersections of protected groups.
\begin{definition} (Statistical Parity Subgroup Fairness)
    Let $\mathcal{G}$ be a collection of protected group indicators $g: A \rightarrow \{0, 1\}$, where $g(s) = 1$ designates that an individual with protected attributes $s$ is in group $g$.  Assume that the mechanism $M(x)$ is binary, i.e. $y \in \{0,1\}$.
    
    Then $M(x)$ is $\gamma$-\emph{statistical parity subgroup fair} (SF) with respect to $\theta$ and $\mathcal{G}$ if for every $g \in \mathcal{G}$, 
    \begin{align}
       & | P_{M,\theta}(M(x) = 1|\theta) - P_{M,\theta}(M(x) = 1) | g(x) = 1, \theta) | \nonumber \\
       & \times P_\theta(g(x) = 1|\theta) \leq \gamma \mbox{ .} \label{eqn:SF}
    \end{align}
\end{definition}
Since we are interested in fairness applications where intersectional ethics are to be upheld, in this work we focus on the case where, similarly to DF, $\mathcal{G}$ contains all possible assignments of the protected attributes $s$ (presumed to be enumerable).  \citet{kearns2018preventing} and \citet{hebert-johnson2018multicalibration} proposed further related multi-attribute definitions regarding false positive rates and calibration, respectively.  Our methods can also be applied to these definitions, but it is beyond the scope of this work. 

\subsection{Empirical Fairness Estimation}
The central challenge for measuring fairness in an intersectional context, either via $\epsilon$-DF, $\gamma$-SF, or related notions, is to estimate $M(x)$'s marginal behavior $P_{M, \theta}(y|s, \theta)$ for each $(y,s)$ pair, with potentially little data for each of these.
The simplest method to do this is to use the empirical data distribution. E.g., for the $\epsilon$-DF criterion, assuming discrete outcomes and protected attributes, $P_{Data}(y|s) = \frac{N_{y,s}}{N_s}$, where $N_{y,s}$ and $N_{s}$ are empirical counts of their subscripted values in the dataset.   \textbf{Empirical differential fairness (EDF)} \citep{foulds2018intersectional} corresponds to verifying that for any $y$, $s_i$, $s_j$,
\begin{equation}
e^{-\epsilon} \leq \frac{N_{y,s_i}}{N_{s_i}}\frac{N_{s_j}}{N_{y,s_j}}\leq e^\epsilon  \mbox{ ,} \label{eqn:discreteDataEDF}
\end{equation}

whenever $N_{s_i} > 0$ and  $N_{s_j} > 0$. 
However, in the intersectional setting, the counts $N_{y,s}$ at the intersection of the values of the protected attributes become rapidly smaller as the dimensionality and cardinality of protected attributes increase (cf. Table \ref{tab:adultIntersectionalCounts}).
In this case, the conditional probabilities in Equations \ref{eqn:DF} and \ref{eqn:SF}, and hence the fairness metrics, will generally have high uncertainty (or variance, from a frequentist perspective) \citep{roth2006modeling}.  The $N_{y,s}$ counts may even be 0, which can make the estimate of $\epsilon$ in Equation \ref{eqn:discreteDataEDF} infinite/undefined.\footnote{Note that \citet{kearns2018preventing} prove large-sample generalization guarantees for empirical estimates of $\gamma$-SF.  As we shall see, this does not imply that empirical estimates of $\gamma$ will be accurate for small-to-moderately sized datasets.  Nevertheless, since SF downweights small groups (the second term of Equation \ref{eqn:SF}) and uses an additive formulation of fairness (compared to DF's multiplicative formulation), it is expected that empirical estimates will be somewhat more stable for SF than DF.}

\section{Model-Based Fairness Estimation}

Instead of using empirical probabilities, in this paper we propose to generalize beyond the training set by learning $P_{M, \theta}(y|s, \theta)$ via a \emph{probabilistic model}.  This approach has several advantages.  First, by exploiting structure in the distributions, e.g. if the mechanism's behavior on \emph{women} is informative of its behavior on \emph{Black women}, we can accurately model all of the conditional probabilities with fewer parameters than empirical frequencies, thereby reducing variance in estimation.  Second, we can use a Bayesian approach to manage uncertainty in the estimation, and to report this uncertainty to an analyst.  

A simple baseline, proposed by \citep{foulds2018intersectional} to address the zero count issue, is to put a Dirichlet prior on the probabilities in Equation \ref{eqn:discreteDataEDF}.  Estimating $\epsilon$-$DF$ via the posterior predictive distribution of the resulting Dirichlet-multinomial, the criterion for any $y$, $s_i$, $s_j$ is
\begin{equation}
e^{-\epsilon} \leq \frac{N_{y,s_i} + \alpha}{N_{s_i}  + |\mathcal{Y}|\alpha}\frac{N_{s_j} + |\mathcal{Y}|\alpha}{N_{y,s_j} + \alpha}\leq e^\epsilon \mbox{ ,} \label{eqn:smoothedFairness}
\end{equation}
where scalar $\alpha$ is each entry of the parameter of a symmetric Dirichlet prior with concentration parameter $|\mathcal{Y}|\alpha$, $\mathcal{Y} = \mbox{Range}(M)$. \citet{foulds2018intersectional} refer to this as \textbf{smoothed EDF}. This can also be used for $\gamma$-SF.


More generally, in this work we propose to estimate $P_{M, \theta}(y|s, \theta)$, and hence the fairness metrics, via \textbf{a probabilistic classifier} that predicts the outcome $y$ given protected attribute values $s\in A$, trained on $\mathcal{D}_s$.  The complexity of the model determines the trade-off between (statistical) bias and variance in the estimation.\footnote{Here, \emph{statistical bias} is not to be confused with unfairness.}  For instance, ordered from high statistical bias to high variance, we could consider \emph{naive Bayes}, \emph{logistic regression}, or \emph{deep neural networks}.

As a compromise between statistical bias and variance in this setting, we also introduce a novel hierarchical extension of logistic regression, where the ``prior'' on $\mbox{logit}(P(y=1|s))$ is a Gaussian around the prediction of a jointly trained logistic regression, allowing deviations justified by sufficient data.  Let $\vec{\mathbf{s}}_j$ be an encoding of protected attribute values $s_j$ with a binary indicator for each attribute's value, with integer $j$ indexing each possible value of $s$, and $\beta_i$ be a regression coefficient for each entry of the $\vec{\mathbf{s}}_j$'s.  The model's generative process is:
\begin{itemize}
    \item     $\sigma_2 \sim \mbox{Exponential}(\lambda)$
    \item $\beta_i \sim \mbox{Normal}(\mu, \sigma_1)$,  \ \ $c \sim \mbox{Normal}(\mu, \sigma_1)$ 
    \item $\gamma_j \sim \mbox{Normal}(\beta^\intercal\vec{\mathbf{s}}_j  + c, \sigma_2)$
    \item $P(y = 1|s_j) = \sigma(\gamma_j))$ ,
\end{itemize}
where $\lambda$ and $\sigma_1$ are prior hyperparameters.  
For most typical models and datasets, to manage uncertainty in the data-sparse intersectional regime, we recommend that the probabilistic classifier be trained via \textbf{fully Bayesian inference}.  
Fully accounting for parameter uncertainty, a single best estimate of the conditional distributions $\hat{\theta}$ to compute $\epsilon$ or $\gamma$ is the posterior predictive distribution, $\hat{\theta} = P_{Model}(y|s, \mathcal{D}_s) = \int_{\bar{\theta}} P_{Model}(y|s, \bar{\theta}) P_{Model}(\bar{\theta}|\mathcal{D}_s)$, for model parameters $\bar{\theta}$.  This can be approximated by, e.g., averaging $P_{Model}(y|s, \bar{\theta})$ over MCMC samples of $\bar{\theta}$ or a variational posterior.  We then report uncertainty in $\epsilon$ by plotting the posterior distribution over $\epsilon$ based on posterior samples of $\bar{\theta}$, and similarly for $\gamma$-SF.  Our overall approach to the Bayesian modeling of intersectional fairness metrics is shown in pseudocode in Algorithm \ref{alg:bayesDF}.  


\section{Bayesian Model Averaging Ensemble}

\begin{algorithm}[t]
\small
\SetAlgoLined
\hspace{-0.1cm}\KwIn{ Development set $\mathcal{D} = \{(x_i, y_i) \}$, mechanism $M(x)$, protected attributes $A$}
\hspace{-0.1cm}\KwOut{ $\hat{\epsilon}_{data}$, $\hat{\epsilon}_{M(x)}$, boxplots of posterior uncertainty in $\epsilon_{data}$, $\epsilon_{M(x)}$, $\epsilon_{M(x)} - \epsilon_{data}$ }
\hspace{-0.1cm}Apply $M(x)$ to $x_i \in \mathcal{D}$, obtain mechanism labels $y_i'$\;
\hspace{-0.1cm}Fit Bayesian classifier $p_1(y|s, \bar{\theta_1})$ on $\mathcal{D}_s = \{(s_i,y_i)\}$\;
\hspace{-0.1cm}Fit Bayesian classifier $p_2(y'|s, \bar{\theta_2})$ on $\mathcal{D}'_s =\{(s_i,y'_i)\}$\;
\hspace{-0.1cm}Estimate $\hat{\epsilon}_{data}$ via Eqn. \ref{eqn:DF} with posterior predictive $p_1(y|s)$\;
\hspace{-0.1cm}Estimate $\hat{\epsilon}_{M(x)}$ via Eqn. \ref{eqn:DF} with posterior predictive $p_2(y'|s)$\;
\hspace{-0.1cm}Plot posterior uncertainty in $\epsilon_{data}$, $\epsilon_{M(x)}$, $\epsilon_{M(x)} - \epsilon_{data}$\;

\caption{
Bayesian estimation of differential fairness and its uncertainty (and similarly for $\gamma$-SF). \label{alg:bayesDF}}
\end{algorithm}

A potential concern with the above approach is that different probabilistic models will lead to different estimates in the measurement of $\epsilon$-DF and $\gamma$-SF.  Consistently with our Bayesian methodology, rather than performing model selection we can account for uncertainty over models by combining them using Bayesian model averaging \citep{hoeting1999bayesian}.  Suppose there are $K$ candidate models.  We estimate the posterior distribution of $\epsilon$ (similarly $\gamma$) in the ensemble given dataset $\mathcal{D}$ via:
\begin{align}
    P(\epsilon|\mathcal{D}) = \sum_{k=1}^K P(\epsilon|M_k, \mathcal{D}) P(M_k| \mathcal{D}) \mbox{ .}
\end{align}
Assuming a uniform prior over models, $P(M_k| \mathcal{D}) \propto \prod_{(y,s) \in \mathcal{D}}P(y| s, M_k)$, the conditional marginal likelihood.  The distribution $P(\epsilon|M_k, \mathcal{D})$ is estimated via MCMC or variational inference over the posterior over the model parameters $P(\bar{\theta}_k|M_k, \mathcal{D})$, with each $\bar{\theta}_k$ corresponding to an $\epsilon$ (or $\gamma$).  Finally, we obtain a gold-standard estimate $\hat{\epsilon}$ or $\hat{\gamma}$ by simulating from the ensemble to estimate the posterior predictive distributions $p(y|s, \mathcal{D})$, and plugging these into Equations \ref{eqn:DF} or \ref{eqn:SF}.

\section{Experimental Results}

%
\begin{table*}[t]
\centering
\small 
\begin{tabular}{c
>{\columncolor[HTML]{ECF4FF}}c 
>{\columncolor[HTML]{ECF4FF}}c cc
>{\columncolor[HTML]{ECF4FF}}c 
>{\columncolor[HTML]{ECF4FF}}c cc}
\multicolumn{9}{c}{Adult Dataset}\\
\hline
                         & \multicolumn{2}{c}{\cellcolor[HTML]{ECF4FF}\begin{tabular}[c]{@{}c@{}}Actual-labeled test set\\ (full training set)\end{tabular}} & \multicolumn{2}{c}{\begin{tabular}[c]{@{}c@{}}$M(x)$-relabeled test set\\ (held-out training subset)\end{tabular}} & \multicolumn{2}{c}{\cellcolor[HTML]{ECF4FF}\begin{tabular}[c]{@{}c@{}}Actual-labeled test set\\ ($10\%$ of the training set)\end{tabular}} & \multicolumn{2}{c}{\begin{tabular}[c]{@{}c@{}}$M(x)$-relabeled test set\\ ($10\%$ of the training subset)\end{tabular}} \\ \cline{2-9} 
\multirow{-2}{*}{Models} & PE                                                               & FB                                                              & PE                                                       & FB                                                      & PE                                                                   & FB                                                                  & PE                                                         & FB                                                         \\ \hline
EDF                      & -0.4366                                                          & -0.4359                                                         & -0.3587                                                  & -0.3580                                                 & -0.4582                                                              & -0.4575                                                             & -0.3959                                                    & -0.3661                                                    \\ \hline
NB                       & -0.4334                                                          & -0.4334                                                         & -0.3646                                                  & -0.3540                                                 & -0.4357                                                              & -0.4478                                                             & -0.3649                                                    & -0.3537                                                    \\ \hline
LR                       & -0.4416                                                          & -0.4304                                                         & -0.3821                                                  & \textbf{-0.3496}                                                 & -0.4533                                                              & -0.4365                                                             & -0.3782                                                    & \textbf{-0.3521}                                                    \\ \hline
DNN                      & -0.4308                                                          & \textbf{-0.4291}                                                         & -0.3645                                                  & -0.3528                                                 & -0.4408                                                              & \textbf{-0.4314}                                                             & -0.3555                                                    & -0.3631                                                    \\ \hline
HLR                      & X                                                                & -0.4323                                                         & X                                                        & -0.3531                                                 & X                                                                    & -0.4384                                                             & X                                                          & -0.3528                                                    \\ \hline
Ensemble                 & \multicolumn{2}{c}{\cellcolor[HTML]{ECF4FF}-0.4337}                                                                                & \multicolumn{2}{c}{-0.3597}                                                                                        & \multicolumn{2}{c}{\cellcolor[HTML]{ECF4FF}-0.4444}                                                                                        & \multicolumn{2}{c}{-0.3647}                                                                                             \\ \hline
\end{tabular}

%
%
\ \\
\begin{tabular}{c
>{\columncolor[HTML]{ECF4FF}}c 
>{\columncolor[HTML]{ECF4FF}}c cc
>{\columncolor[HTML]{ECF4FF}}c 
>{\columncolor[HTML]{ECF4FF}}c cc}
\multicolumn{9}{c}{COMPAS Dataset}\\
\hline
                         & \multicolumn{2}{c}{\cellcolor[HTML]{ECF4FF}\begin{tabular}[c]{@{}c@{}}Actual-labeled test set\\ (full training set)\end{tabular}} & \multicolumn{2}{c}{\begin{tabular}[c]{@{}c@{}}$M(x)$-relabeled test set\\ (held-out training subset)\end{tabular}} & \multicolumn{2}{c}{\cellcolor[HTML]{ECF4FF}\begin{tabular}[c]{@{}c@{}}Actual-labeled test set\\ ($10\%$ of the training set)\end{tabular}} & \multicolumn{2}{c}{\begin{tabular}[c]{@{}c@{}}$M(x)$-relabeled test set\\ ($10\%$ of the training subset)\end{tabular}} \\ \cline{2-9} 
\multirow{-2}{*}{Models} & PE                                                               & FB                                                              & PE                                                       & FB                                                      & PE                                                                   & FB                                                                  & PE                                                         & FB                                                         \\ \hline
EDF                      & -0.6899                                                          & -0.6889                                                         & -0.6491                                                  & -0.6490                                                  & -0.6745                                                              & -0.6736                                                             & -0.6652                                                    & -0.6610                                                     \\ \hline
NB                       & -0.6882                                                          & -0.6851                                                         & -0.6456                                                  & -0.6447                                                 & -0.6748                                                              & -0.6818                                                             & -0.6517                                                    & \textbf{-0.6510}                                                     \\ \hline
LR                       & -0.6926                                                          & -0.6835                                                         & -0.6469                                                  & \textbf{-0.6436}                                                 & -0.6862                                                              & -0.6812                                                             & -0.6579                                                    & -0.6538                                                    \\ \hline
DNN                      & -0.6725                                                          & \textbf{-0.6721}                                                         & -0.6461                                                  & -0.6593                                                 & -0.6721                                                              & \textbf{-0.6714}                                                             & -0.6535                                                    & -0.6566                                                    \\ \hline
HLR                      & X                                                                & -0.6859                                                         & X                                                        & -0.6457                                                 & X                                                                    & -0.6721                                                             & X                                                          & -0.6569                                                    \\ \hline
Ensemble                 & \multicolumn{2}{c}{\cellcolor[HTML]{ECF4FF}-0.6843}                                                                                & \multicolumn{2}{c}{-0.6478}                                                                                        & \multicolumn{2}{c}{\cellcolor[HTML]{ECF4FF}-0.6764}                                                                                        & \multicolumn{2}{c}{-0.6564}                                                                                             \\ \hline
\end{tabular}
\caption{Comparison of predictive performance of intersectional fairness models with respect to average negative cross-entropy per intersection on the test set (higher is better), on Adult (top) and COMPAS (bottom). Here, PE = point estimate, FB = fully Bayesian estimate using the posterior predictive distribution.   EDF-FB is the Dirichlet-multinomial model, cf. Equation \ref{eqn:smoothedFairness}.  The best performing method is indicated in bold.}
\label{table:both-cross}
\end{table*}

%
%

The goals of our experiments were to compare our proposed Bayesian modeling approach for estimating intersectional fairness to point estimation and to empirical measurement, to evaluate the performance of different models and of model averaging, to study the effect of uncertainty/variance in intersectional fairness estimation, and to illustrate the practical application of our methods.
We performed all experiments on two datasets:
\begin{itemize}
    \item The Adult 1994 U.S. census income data from the UCI repository \citep{kohavi-nbtree}.  This dataset consists of $14$ attributes regarding work, relationships, and demographics for individuals, who are labeled according to whether their income exceeds $\$50,000$ per year, pre-split into a training set of $32,561$ instances and a test set of $16,281$ instances. We select race, gender, and nationality as the protected attributes. As most instances have U.S. nationality, we treat nationality as binary between U.S. and ``other.'' Gender is also coded as binary. The race attribute originally had 5 values. We merged the Native American category with ``other,'' as both contained very few instances.\footnote{The decision to merge attribute values was made for a previous study (and similarly for the COMPAS dataset below).  Leaving these values unmerged would likely have increased the relative benefit of our methods.  } 
    \item The COMPAS dataset regarding a system that is used to predict criminal recidivism, and which has been criticized as potentially biased ~\citep{angwin2016machine}. We used race and gender as protected attributes. Gender was coded as binary. Race originally had 6 values, but we merged ``Asian'' and ``Native American'' with ``other,'' as all three contained very few instances.  We used ``actual recidivism'' (within a 2-year period), which is binary, as the true label of the data generating process and the COMPAS system's prediction as the labels from $M(x)$. 
    Following \cite{angwin2016machine}, we merged the ``medium'' and ``high'' labels to make COMPAS scores binary, since the actual labels are binary.  For evaluating our models, we split the COMPAS dataset into train and test sets with 5,410 and 1,804 data instances, respectively.
\end{itemize}
All models were trained using PyMC3, with ADVI used for Bayesian inference.  Posterior predictive distributions were estimated by sampling from the variational posterior and averaging the predictions.  For ease of reading, important observations are indicated in bold. 

\subsection{Prediction Performance on Held-Out Data}

\begin{figure*}     
        \begin{minipage}{0.5\textwidth}
        \centering Differential Fairness
        \end{minipage}
        \begin{minipage}{0.5\textwidth}
        \centering Subgroup Fairness
        \end{minipage}
		\centerline{\includegraphics[width=0.51\textwidth]{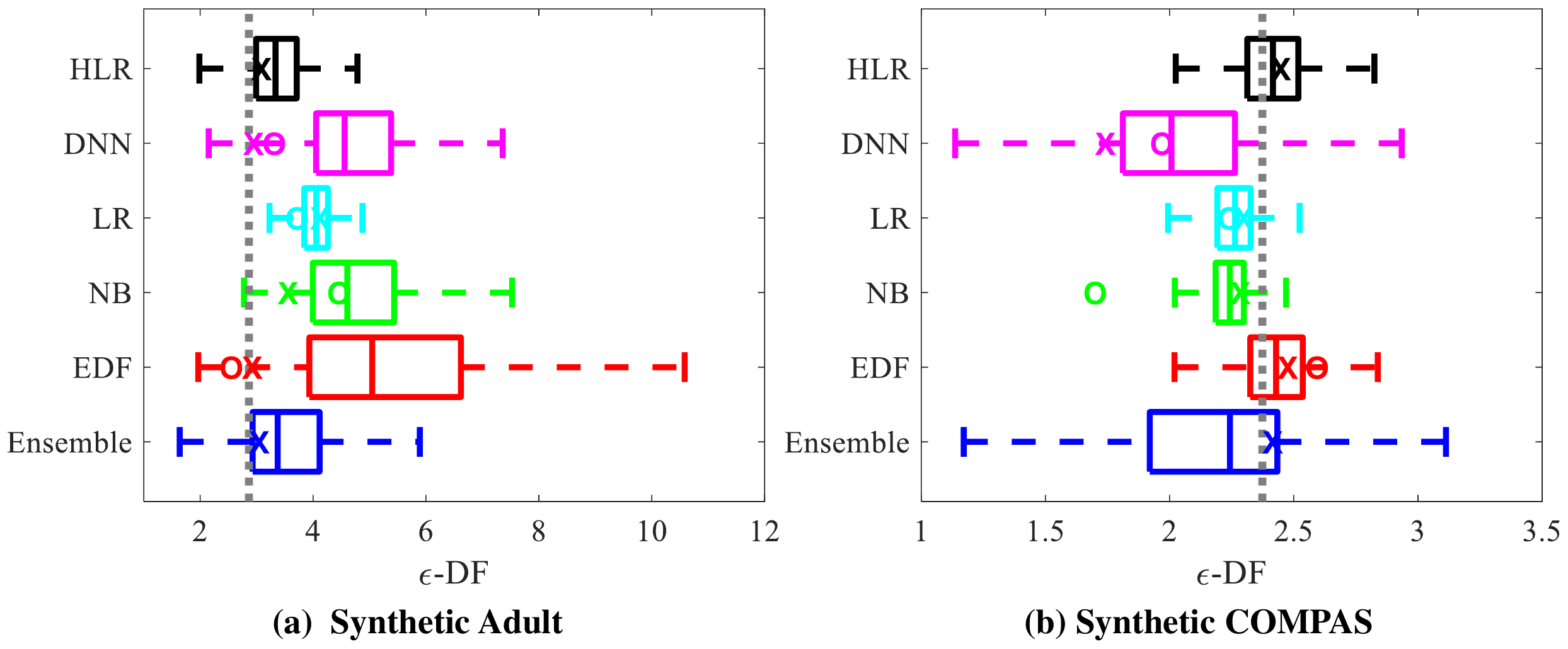} \includegraphics[width=0.51\textwidth]{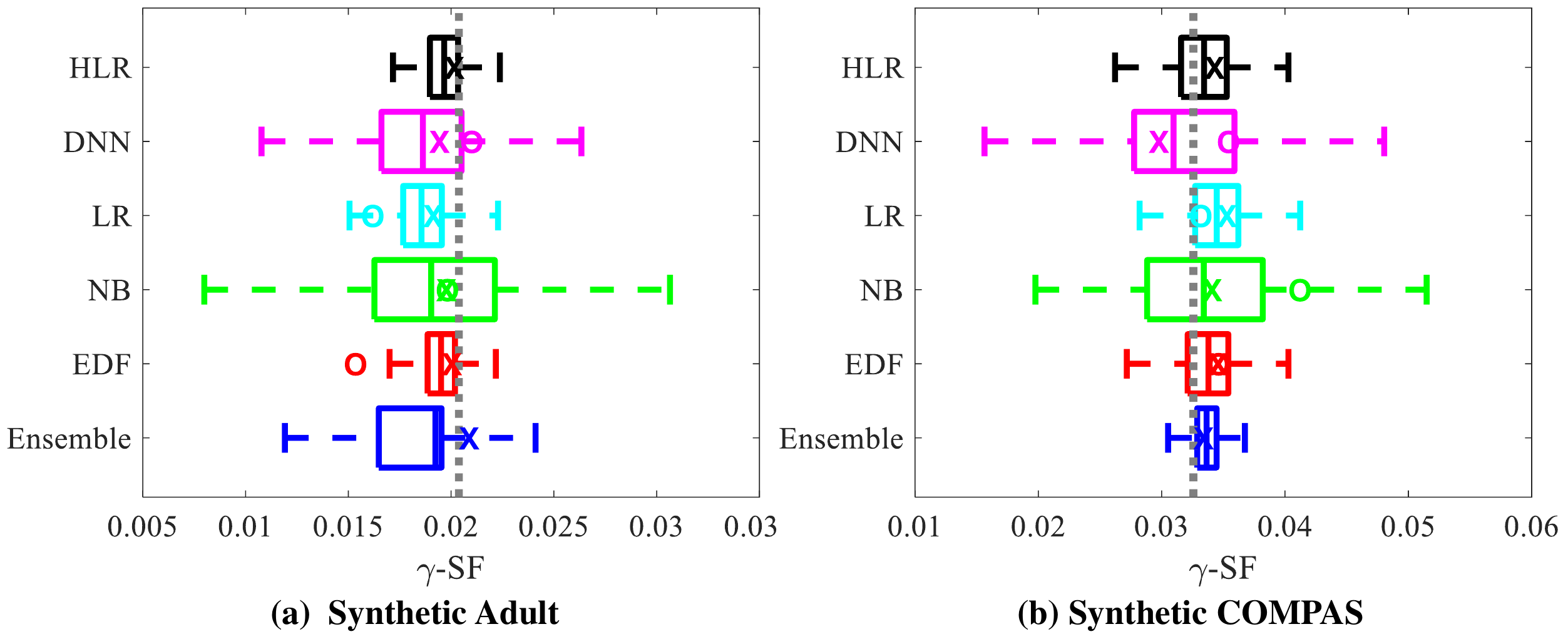}}
%
		\caption{\small 
		Fairness estimates using variational posteriors, point estimates, and posterior predictive distributions: semi-synthetic versions of (a) Adult and (b) COMPAS datasets, for differential fairness (left) and subgroup fairness (right). The ``O'' and ``X'' on top of the box-plots indicate estimates from PE and the posterior predictive distribution of FB models, respectively. The dotted vertical line represents ground truth $\epsilon$ or $\gamma$. The data were generated using per-group Gaussians and a threshold decision boundary.
		}
		\label{fig:syntheticDfandSF} 
\end{figure*}
%

We first studied the predictive performance for models of $P_{M, \theta}(y|s, \theta)$, as needed to compute $\epsilon$-DF and $\gamma$-SF:  the empirical distribution (EDF), naive Bayes (NB), logistic regression (LR), deep neural networks (DNN), and our hierarchical logistic regression model (HLR). For each model, we compare point estimates (PE) (MAP, except for EDF), and fully Bayesian inference via the posterior predictive distribution (FB), as well as a Bayesian model averaging ensemble (Ensemble) of all PE and FB models. Note that the configuration of the DNN architecture is $3$ hidden layers, $10$ neurons in each layer, ``relu'' and ``sigmoid'' activations for the hidden and output layers, respectively. The Dirichlet multinomial model (cf. Eqn. \ref{eqn:smoothedFairness}) is denoted EDF-FB.  
%
We trained all models on the training set and reported negative cross-entropy from the test set's empirical $P(y|s)$ and $P(y'|s)$, averaged over intersections $s$.  Negative cross-entropy is closely related to log-likelihood, and here measures the similarity of the model's conditional distributions to those of the test set.

Results on the Adult and COMPAS datasets are shown in Table \ref{table:both-cross}.  We report results both for $y$ labels in the test data (inequity in society), and for an algorithmic mechanism $y' = M(x)$.  
For Adult, we set $M(x)$ to be a logistic regression model, since it has an appropriate level of model complexity for this data regime. We trained the model on half of the training set (which was held out from the $P_{M, \theta}(y|s, \theta)$ models). For COMPAS, the mechanism $M(x)$ is the COMPAS system itself.  Although COMPAS is a black box, we observe its assigned class labels $y'$, and our models extrapolate its behavior on intersectional groups.
Furthermore, to simulate a scenario with more sparse data,  we repeated these experiments using only $10\%$ of the training data (for Adult, 6,512 instances for actual labels, and 3,256 instances for $M(x)$ labels).

We found that in the vast majority of cases, the \textbf{probabilistic models outperformed empirical estimates} EDF-PE and EDF-FB in terms of prediction performance, and that \textbf{fully Bayesian inference outperformed point estimates}.  The best method was a fully Bayesian model in all cases.  These differences were typically \emph{magnified in the more-sparse regime} where only 10\% of the data was used.     Deep neural networks (DNNs) were the best predictor for the actual labels in the data, i.e. bias in society, but performed worse at predicting the behavior of algorithms $M(x)$, where in some cases they failed to outperform the EDF baselines.  When predicting the behavior of algorithms, logistic regression and naive Bayes were the best performing methods.  We hypothesize that these differences are primarily because real-world class distributions are more complex than $M(x)$'s distributions, and partly because more data was available for the actual-label scenario due to the hold-out procedure. 

While our Bayesian hierarchical logistic regression (\textbf{HLR-FB}) method was never the best predictor on any one dataset, it exhibited the \textbf{most reliable behavior across data sets}, being the only method to outperform the empirical distribution (EDF-PE) and the Dirichlet-multinomial baseline (EDF-FB) in all cases.  The point estimate (PE) version of HLR performed too poorly to be shown. This may be due to numerical instability in PyMC3.  The Bayesian model average (\textbf{Ensemble}) was also relatively \textbf{stable across datasets}, but HLR outperformed it here in most cases.

\subsection{Fairness Metrics on Semi-Synthetic Data}
\begin{figure*}[t]
        \begin{minipage}{\textwidth}
            \centering Differential Fairness
        \end{minipage}
		\centerline{\includegraphics[width=0.85\textwidth]{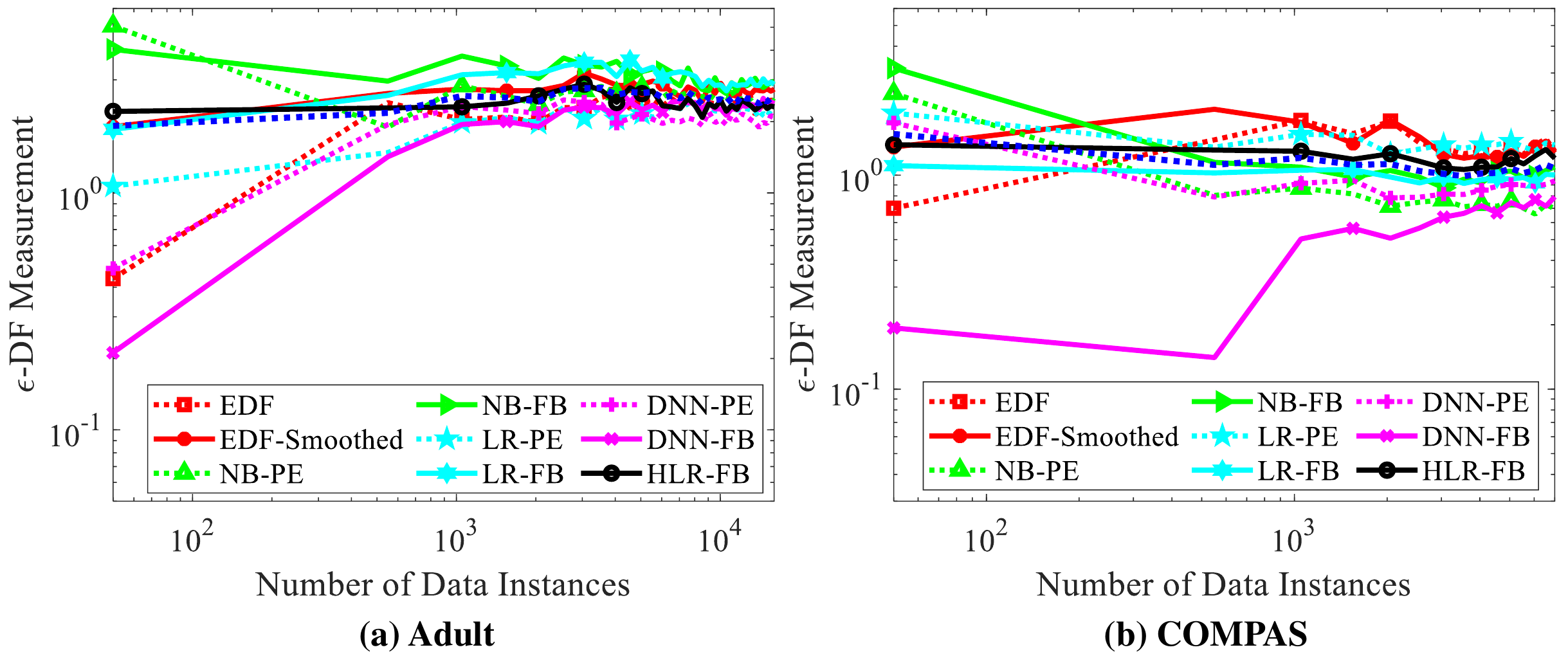}}
        \begin{minipage}{\textwidth}
            \centering Subgroup Fairness
        \end{minipage}
		\centerline{\includegraphics[width=0.85\textwidth]{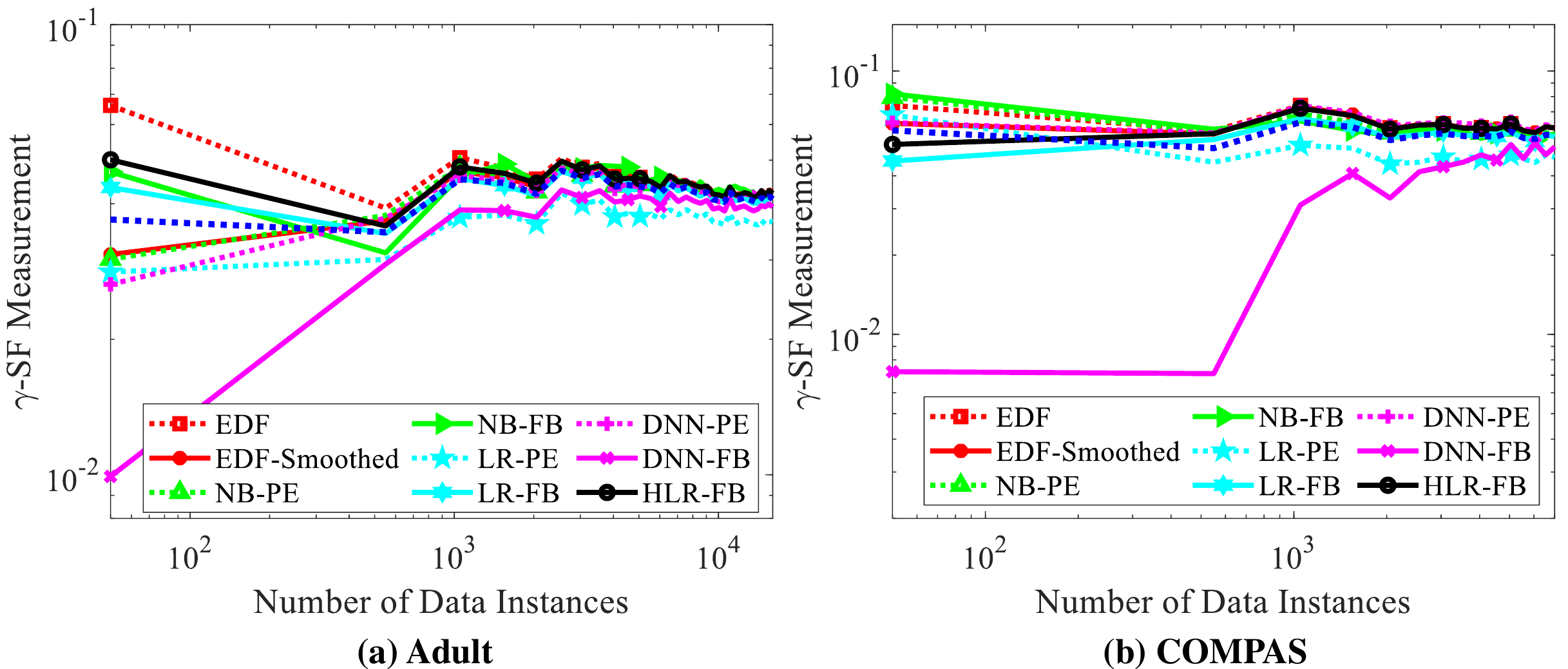}}
		\caption{\small $\epsilon$-DF (top) and $\gamma$-SF (bottom) measurement of algorithm $M(x)$ for (a) logistic regression on the Adult dataset and (b) the COMPAS algorithm, using different $P_{M, \theta}(y|s, \theta)$ models, versus number of instances, averaged over 10 bootstrap samples. The dotted blue line indicates the Bayesian ensemble approach.}
		\label{fig:epsilonAndGammaPerInstance} 
\end{figure*}

%

In this section, we compare all the models with respect to the deviation of their fairness estimates from the ground truth. Since we cannot compute ground truth fairness metrics without knowing the true data distribution $\theta$, we design these experiments on semi-synthetic versions of the Adult and COMPAS datasets. We use the same number of protected attribute values, and instances per intersectional group as for COMPAS and Adult test datasets, but where the class probabilities are determined by a Gaussian model with a threshold decision boundary. 

In our Gaussian threshold model, suppose that $x$ is a ``risk score'' encoding the untrustworthiness of an individual, generated from a Gaussian given the individual's protected attributes $s$.  The mechanism $M(x) = x \geq t$ assign a ``high risk of recidivism'' label if the individual's risk score exceeds threshold $t$. We generate the binary class labels for our semi-synthetic COMPAS and Adult datasets by drawing the same number of instances $x$ per intersectional group as in the original data, and assigning class labels using $M(x)$.  We generate the data via $P(x|s) = N(x;\mu=w_{s}\times \sum_{d}s_{d},\sigma=1)$, where $s_{d}$ is the $d$th protected attribute value for the individual encoded as an integer, $w_{s} \in (0,1)$ is a group-specific weight, and $t = 2.5$. 
We chose $w_{s} = P_{data}(y=1|s)$ plus a small constant, thereby making the synthetic data $P(y=1|s)$'s have some association with the empirical $P_{data}(y=1|s)$'s.  The overall process creates semi-synthetic data with correlations between intersectional groups, and reasonable ground truth values of $\epsilon$ and $\gamma$.

Figure~\ref{fig:syntheticDfandSF} shows DF and SF estimates for all models on both semi-synthetic datasets, with the gray dotted vertical lines on top of the plots indicating the ground truth $\epsilon$-DF and $\gamma$-SF. Fully Bayesian inference allow us to encode uncertainty in the fairness metrics (box-plots) as well as a ``best'' estimate using the posterior predictive distribution (``$X$'').  Point estimates (PE) via MAP or the empirical distribution (for EDF), are indicated as ``$O$''.

Our first finding is that although \textbf{empirical estimates} of $\epsilon$ and $\gamma$ (the ``$O$'' for EDF) are in some cases accurate, in others they can \textbf{deviate substantially from the true values}.  The \textbf{Bayesian estimates} of $\epsilon$ and $\gamma$, using the posterior predictive (``$X$''), were \textbf{closer to the ground truth} compared to PE (``$O$'') methods.  Note that the posterior predictive estimates ``$X$'', calculated via an average over $p(y|s,\bar{\theta})$'s to compute a single $\epsilon$, often deviated substantially from the \textbf{posterior median} $\epsilon$, calculated as an average over the $\epsilon$'s corresponding to each posterior sample of $\bar{\theta}$, and which was sometimes quite \textbf{far from the ground truth} (and similarly for $\gamma$).

Our \textbf{HLR and Bayesian Ensemble approaches performed the best}, in that their fairness estimates from the posterior predictive were overall the closest to the ground truth for both $\epsilon$-DF and $\gamma$-SF.  The Ensemble typically reported higher posterior variance than HLR, due to averaging over multiple models.  The EDF-FB Dirichlet-multinomial model (boxplot and ``$X$'' for EDF) had accurate posterior predictive estimates, obtained using Equation \ref{eqn:smoothedFairness} for $\epsilon$, and similarly for $\gamma$. However, on the Synthetic Adult dataset the EDF-FB posterior distribution over $\epsilon$ failed to capture the ground truth value, unlike HLR and the Bayesian Ensemble. 

\subsection{Stability of Estimation vs Data Sparsity}

\begin{figure*}[t]     
		\centerline{\includegraphics[width=0.85\textwidth]{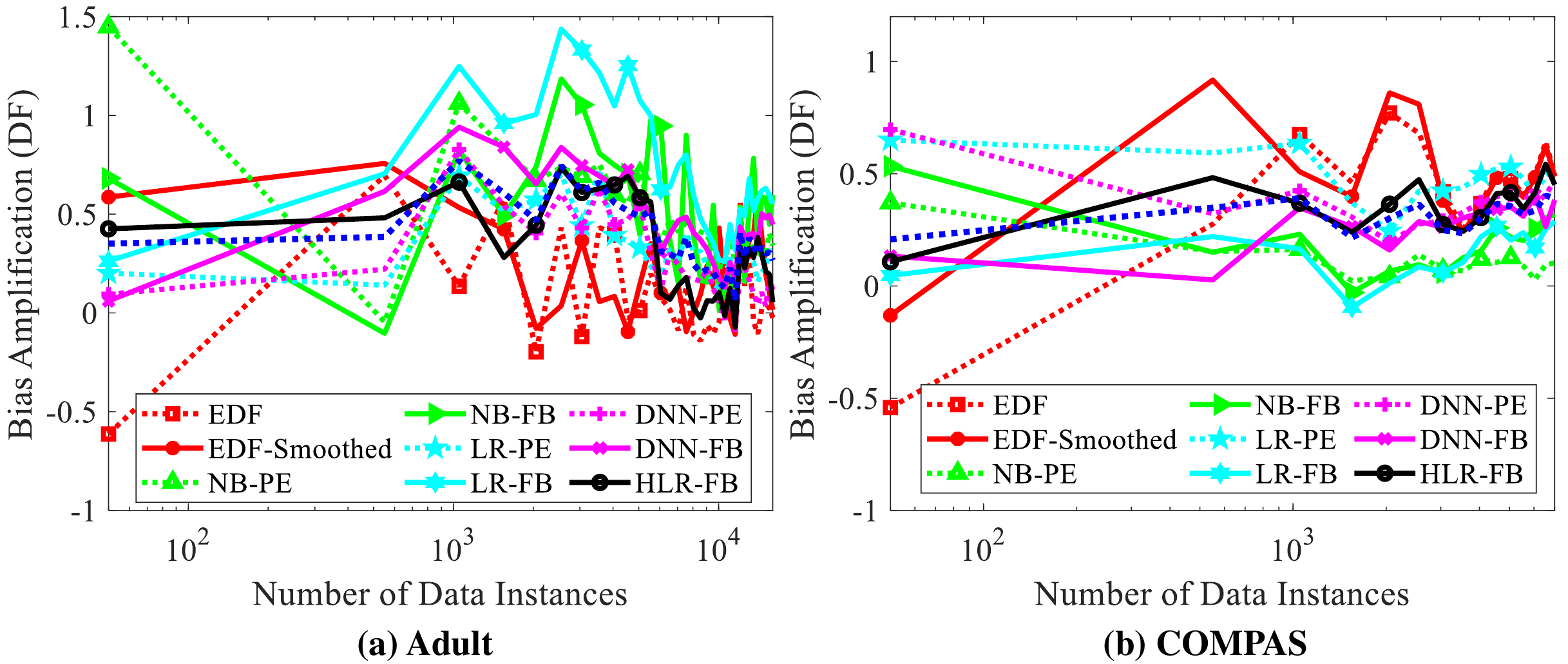}}
		\caption{\small $(\epsilon_2 - \epsilon_1)$-$DF$ bias amplification measurement of algorithm $M(x)$ for (a) logistic regression on the Adult dataset and (b) the COMPAS algorithm, using different $P_{M, \theta}(y|s, \theta)$ models, versus number of data instances, averaged over 10 bootstrap samples. The dotted blue line indicates Bayesian ensemble approach.}
		\label{fig:biasDFPerInstance} 
\end{figure*}
We now turn to the study of intersectional fairness estimation on the real datasets.  We first investigated the stability of the estimation of the fairness metrics versus data sparsity, by estimating $\epsilon$ from bootstrap samples of the datasets, varying the number of samples (Figure~\ref{fig:epsilonAndGammaPerInstance}).  
For each number of data instances, we  generated $10$ bootstrap datasets and reported the average $\epsilon$-DF and $\gamma$-SF for each model on the Adult and COMPAS datasets. 

For both fairness metrics, the estimates differed greatly between models in the small-data regime, and the models converged to relatively similar estimates as the amount of data increased.  The empirical (EDF) estimates were often very noisy with little data, compared to most other models.  Except for deep neural networks (DNN), \textbf{Bayesian models} (solid lines) were typically found to \textbf{converge more quickly in the amount of data} to the consensus full-data estimates, compared to point estimates (dashed lines). The Bayesian DNN (DNN-FB)'s estimates of $\epsilon$ deviated substantially from the full data estimates in the low data regime, likely indicating poor performance. This may be due to the overparameterization of the model, and/or convergence issues.  

The proposed \textbf{HLR-FB} model was relatively \textbf{stable in the number of instances}, and produced estimates of the fairness metrics which were similar to all models' full-data estimates, even when the number of instances was very small.  The Ensemble also exhibited this behavior.

To analyze the models' performance in the very sparse data regime in more detail, we compared their small data fairness metric estimates, calculated at the left end of the curves in Figure \ref{fig:epsilonAndGammaPerInstance} ($1\%$ of the data), with the full data ``ground truth'' (the right end of the curves).  We approximate the ground truth $\epsilon$ and $\gamma$ as the median of all bootstrap samples for all the models, where the size of the bootstrap samples is the size of the full dataset, and we report the average L1 distance from the small data estimates and the approximate ground truth (Table~\ref{table:deviation}).  We found that \textbf{in the sparse data regime}, the \textbf{fully Bayesian models} (\textbf{FB}) had \textbf{smaller deviations} from the full data ``\textbf{ground truth}'' estimates. Our \textbf{HLR-FB model and logistic regression performed the best} in this sparse data regime including the Bayesian model averaging Ensemble method.   

In Figure \ref{fig:biasDFPerInstance}, we further studied the impact of dataset size on the \textbf{DF bias amplification metric} (Definition \ref{def:biasAmplification}).  Since this is calculated as the difference of two noisy estimates of $\epsilon$-DF, the relative noise was higher.  The methods differed in the estimated direction of the bias amplification (increase or decrease) in the small data regime, but all pointed to a positive increase with the full data, on both datasets.  The overall conclusions were similar to the previous experiments, and the \textbf{HLR-FB} and \textbf{Ensemble} methods were once again the \textbf{most stable when given little data}.  We report results on a bias amplification version of $\gamma$-SF in the Appendix.  Note that in these experiments, averaging over bootstraps improves the stability of high variance estimation methods, and the estimates may differ in individual bootstrap samples (we report examples in the Appendix).  Setting aside pure Bayesian or frequentist ideologies, to estimate the fairness metrics in practice it may be useful to \textbf{use bootstrap averaging in conjunction with Bayesian models}, as performed here. 
%

\subsection{Case Study on COMPAS Dataset}

\begin{figure*}
        \begin{minipage}{\textwidth}
            \centering Differential Fairness
        \end{minipage}
		\centerline{\includegraphics[width=0.9\textwidth]{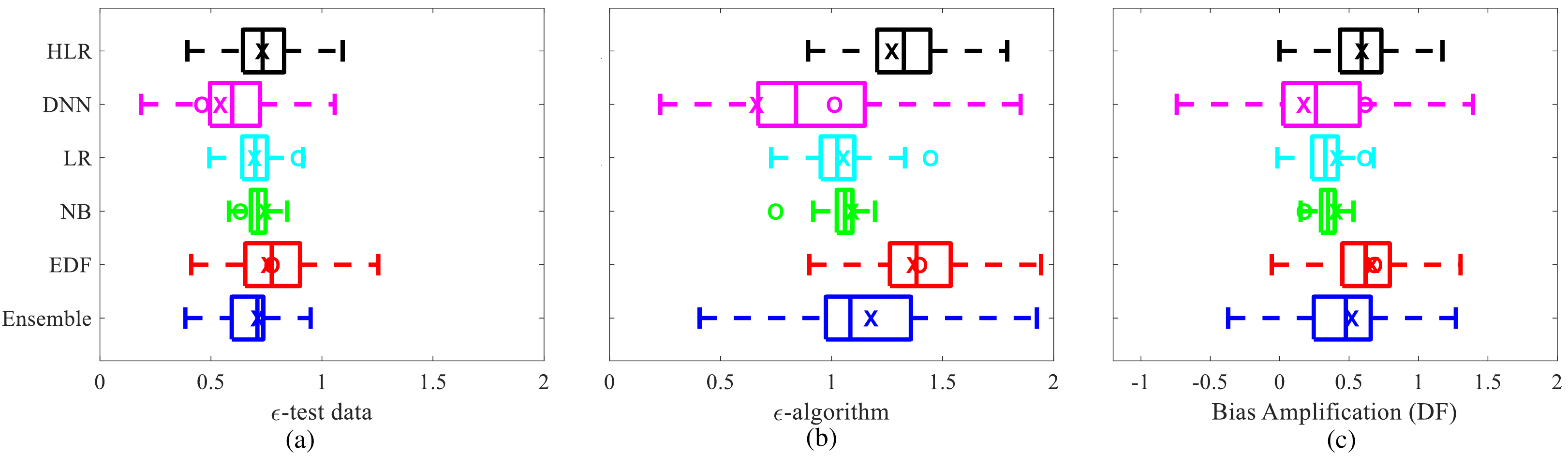}}
        \begin{minipage}{\textwidth}
            \centering Subgroup Fairness
        \end{minipage}
		\centerline{\includegraphics[width=0.9\textwidth]{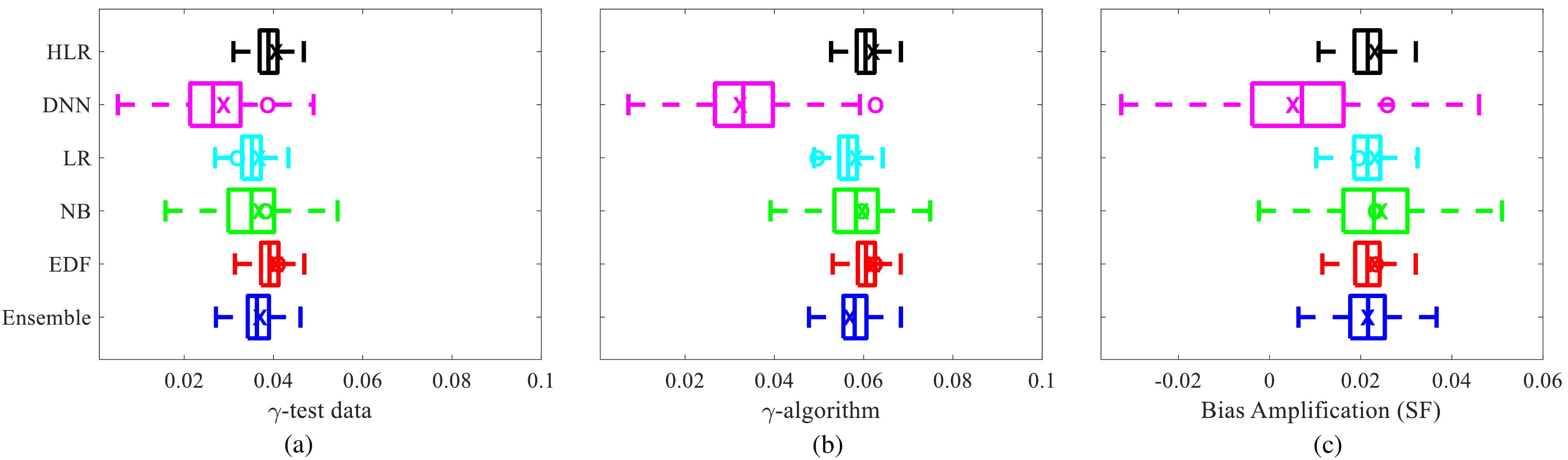}}
		\caption{\small $\epsilon$-DF (top) and $\gamma$-SF (bottom) estimates using PE and FB variational posterior distributions, on the COMPAS dataset: (a) fairness estimates on true recidivism label of data, (b) fairness estimates on COMPAS system $M(x)$-relabeled data, and (c) bias amplification by the COMPAS system, for both DF $(\epsilon_{(b)} - \epsilon_{(a)})$ and SF $(\gamma_{(b)} - \gamma_{(a)})$. The ``O'' and ``X'' on top of the box-plots indicates estimates from PE models and the posterior predictive distribution of FB models, respectively.}
		\label{fig:df-sf-box-compas} 
\end{figure*}

\begin{table}[t]
\centering
\resizebox{0.49\textwidth}{!}{
\begin{tabular}{ccccccccc}
\hline
                        & \multicolumn{4}{c}{$1\%$ of Adult Dataset}                                                      & \multicolumn{4}{c}{$1\%$ of COMPAS Dataset}                                                     \\ \cline{2-9} 
                        & \multicolumn{2}{c}{\cellcolor[HTML]{ECF4FF}$\epsilon$-DF}     & \multicolumn{2}{c}{$\gamma$-SF} & \multicolumn{2}{c}{\cellcolor[HTML]{ECF4FF}$\epsilon$-DF}     & \multicolumn{2}{c}{$\gamma$-SF} \\ \cline{2-9} 
\multirow{-3}{*}{Models} & \cellcolor[HTML]{ECF4FF}PE    & \cellcolor[HTML]{ECF4FF}FB    & PE             & FB             & \cellcolor[HTML]{ECF4FF}PE    & \cellcolor[HTML]{ECF4FF}FB    & PE             & FB             \\ \hline
EDF                     & \cellcolor[HTML]{ECF4FF}2.105 & \cellcolor[HTML]{ECF4FF}0.740 & 0.028          & 0.019          & \cellcolor[HTML]{ECF4FF}0.541 & \cellcolor[HTML]{ECF4FF}0.485 & 0.028          & 0.022          \\ \hline
NB                      & \cellcolor[HTML]{ECF4FF}2.644 & \cellcolor[HTML]{ECF4FF}1.614 & 0.024          & 0.017          & \cellcolor[HTML]{ECF4FF}1.475 & \cellcolor[HTML]{ECF4FF}2.083 & 0.031          & 0.031          \\ \hline
LR                      & \cellcolor[HTML]{ECF4FF}1.367 & \cellcolor[HTML]{ECF4FF}0.572 & 0.019          & \textbf{0.008}          & \cellcolor[HTML]{ECF4FF}0.901 & \cellcolor[HTML]{ECF4FF}\textbf{0.208} & 0.021          & 0.016          \\ \hline
DNN                     & \cellcolor[HTML]{ECF4FF}1.958 & \cellcolor[HTML]{ECF4FF}2.210 & 0.016          & 0.031          & \cellcolor[HTML]{ECF4FF}0.692 & \cellcolor[HTML]{ECF4FF}0.884 & 0.022          & 0.051          \\ \hline
HLR                     & \cellcolor[HTML]{ECF4FF}X     & \cellcolor[HTML]{ECF4FF}\textbf{0.341} & X              & 0.011          & \cellcolor[HTML]{ECF4FF}X     & \cellcolor[HTML]{ECF4FF}0.393 & X              & \textbf{0.015}          \\ \hline
Ensemble                & \multicolumn{2}{c}{\cellcolor[HTML]{ECF4FF}1.489}             & \multicolumn{2}{c}{0.019}       & \multicolumn{2}{c}{\cellcolor[HTML]{ECF4FF}0.835}             & \multicolumn{2}{c}{0.026}       \\ \hline
\end{tabular}
}
\caption{L1 Deviations of $\epsilon$-DF and $\gamma$-SF measurements with $1\%$ of Adult and COMPAS dataset from full dataset ``ground truth'' estimates, to show the effect of data sparsity (lower is better). FB methods perform better than PE methods. Our HLR-FB model often performs the best.} 
\label{table:deviation}
\end{table}
%

As a practical case study, we estimated the intersectional fairness metrics, and their uncertainty via the variational posteriors, on the COMPAS dataset (Figure~\ref{fig:df-sf-box-compas}).  All models place high posterior probability on substantially high unfairness values $\epsilon$ and $\gamma$, and they indicate that the direction of bias amplification is almost certainly positive for both metrics.  
To interpret $\epsilon$-DF, note that the 80\% rule, used as a legal standard for evidence of disparate impact discrimination \citep{eeoc1966guidelines}, finds evidence of discrimination if $\epsilon \geq -\log 0.8 = 0.2231$.\footnote{DF calculates ratios of probabilities for all $y$.  Strictly, the 80\% rule is calculated on the favorable outcome only.}  All models put most of their posterior density, and their posterior predictive estimates, on values higher than this for true recidivism, COMPAS, and its bias amplification. The most reliable model, HLR-FB, predicts that the DF bias amplification of COMPAS is most likely around $0.5$-DF, with lower and upper posterior quartiles at around $0.35$ and $0.65$, respectively.  The results show strong evidence that COMPAS increases the bias beyond the inequities in the data.  We report a similar analysis on Adult in the Appendix.

\section{Discussion: Practical Recommendations}
We showed that fully Bayesian models provide more reliable estimates of intersectional fairness metrics than empirical estimates and point estimates.  Although the best model depends on the data regime, our proposed HLR-FB model provides stable estimates compared to other methods, particularly in the very sparse data setting.  We found that a Bayesian model averaging ensemble also improves stability in estimation, but it did not outperform HLR-FB on its own.  We therefore recommend the use of HLR-FB as a reliable intersectional fairness estimation method with sparse multi-attribute data.

\section{Conclusion}
\label{sec:conclusion}
We have proposed Bayesian modeling approaches to reliably estimate fairness and its uncertainty in the sparse data regime which arises from multi-attribute intersectional fairness definitions.  
Our empirical results show the benefits of the probabilistic model-based approach in this setting compared to empirical probability estimates, especially when using Bayesian inference.
We proposed a Bayesian hierarchical logistic regression model which provides stable estimates of fairness metrics with sparse intersectional data, and we applied our methods to study the bias in the COMPAS recidivism predictor and a model trained on census data.
We plan to develop extensions to model continuous protected attributes, more sophisticated latent variable modeling approaches, and learning algorithms which incorporate uncertainty in fairness measurement during training.

\newpage

\section*{Acknowledgments}
This work was performed under the following financial assistance award: 60NANB18D227 from U.S. Department of Commerce, National Institute of Standards and Technology.

We thank Rosie Kar for valuable advice and feedback regarding intersectional feminism.

\bibliographystyle{named}
\bibliography{references}

\clearpage
\appendix

\section{Appendix: Additional Experimental Results}
\begin{figure*}[t]
        \begin{minipage}{\textwidth}
            \centering Differential Fairness
        \end{minipage}
		\centerline{\includegraphics[width=0.85\textwidth]{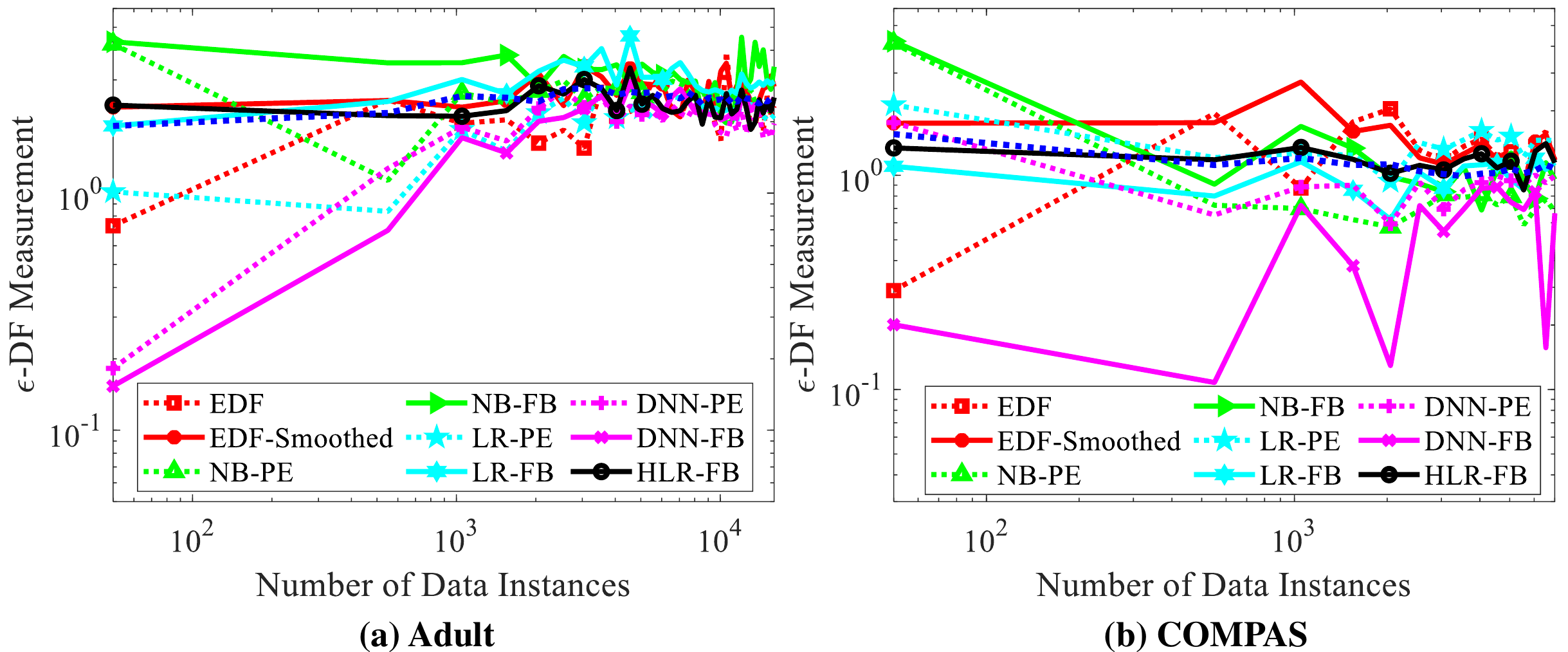}}
        \begin{minipage}{\textwidth}
            \centering Subgroup Fairness
        \end{minipage}
		\centerline{\includegraphics[width=0.85\textwidth]{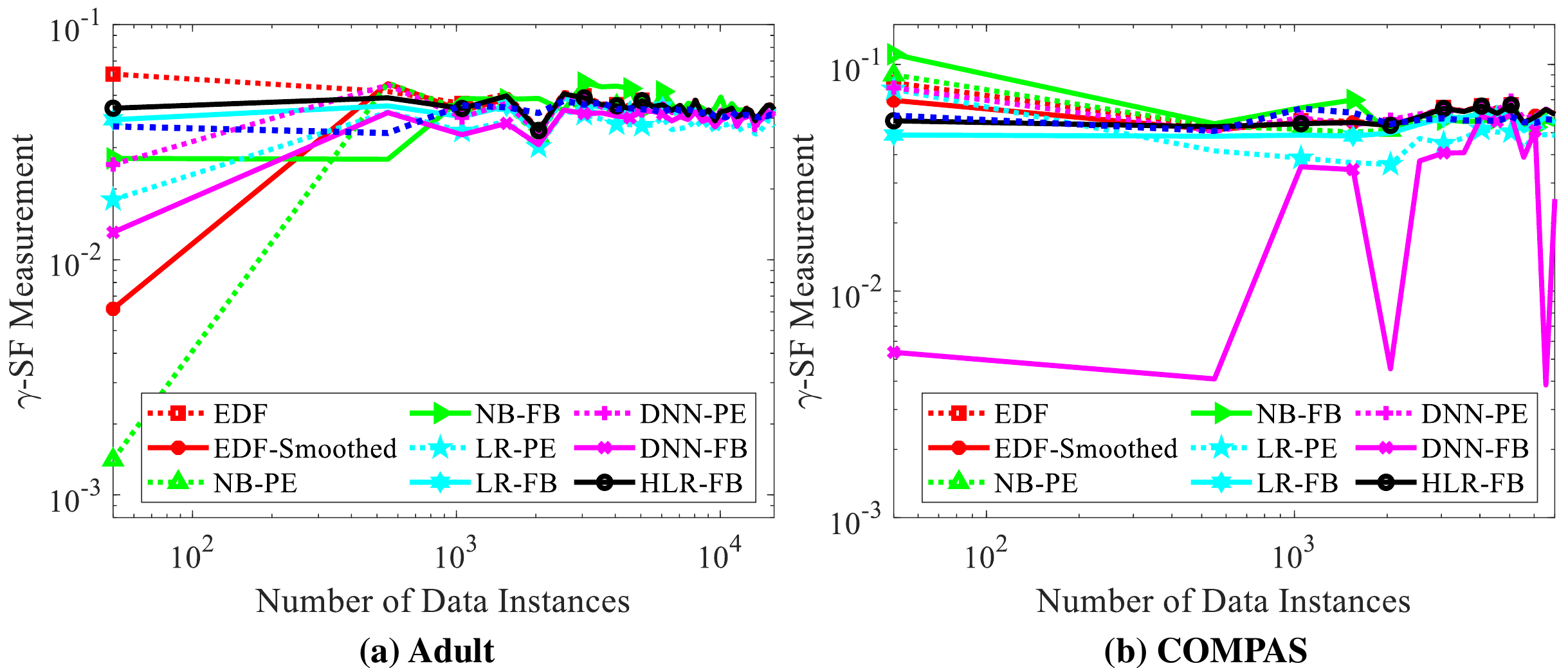}}
		\caption{\small $\epsilon$-DF (top) and $\gamma$-SF (bottom) measurement of an algorithm $M(x)$ for (a) logistic regression on the Adult dataset and (b) the COMPAS algorithm, using different $P_{M, \theta}(y|s, \theta)$ models, versus the number of instances, for a randomly chosen bootstrap data sample. For a reference to compare to the other models, we report the average over 10 bootstrap samples for the Bayesian ensemble approach, rather than using a single bootstrap sample, as for the other methods (dotted blue line).}
		\label{fig:RandomepsilonAndGammaPerInstance} 
\end{figure*}
\begin{figure*}[t]     
		\centerline{\includegraphics[width=0.85\textwidth]{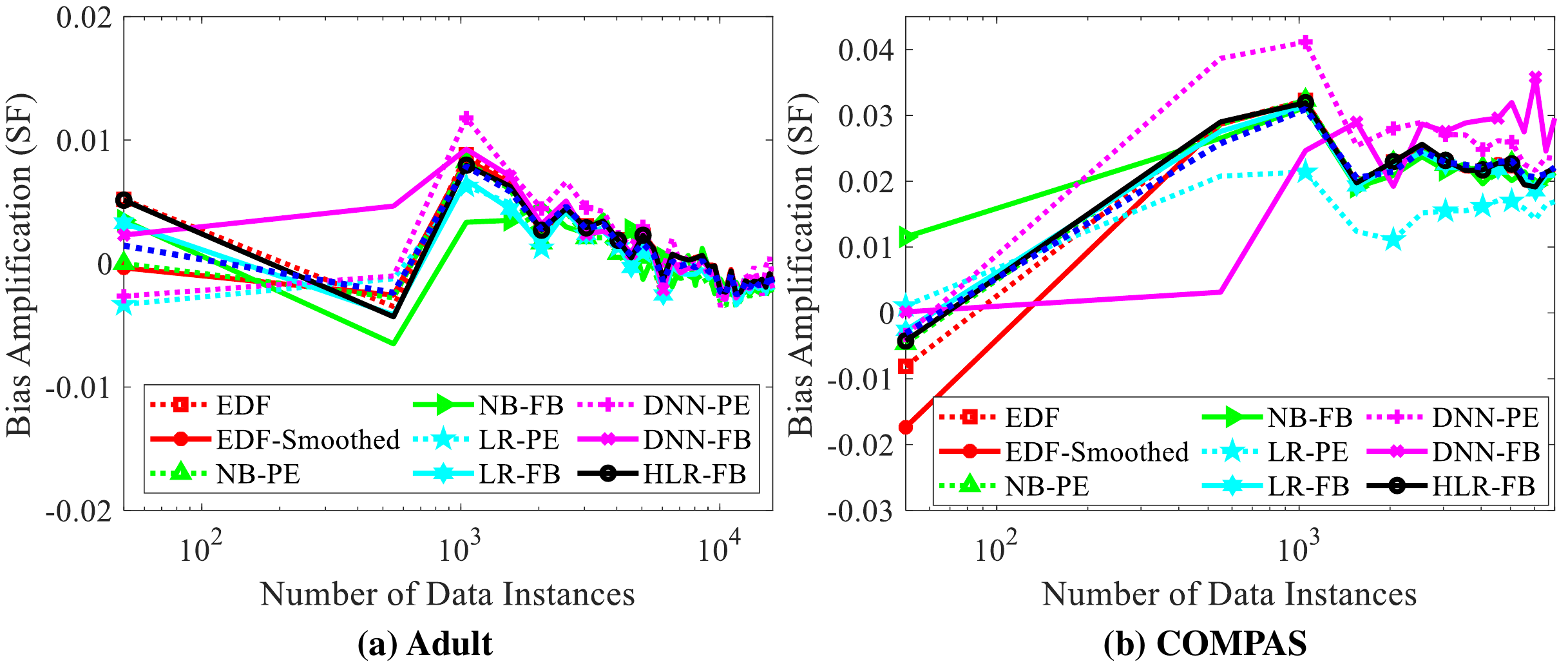}}
		\caption{\small $(\gamma_2 - \gamma_1)$-$SF$ bias amplification by the mechanism $M(x)$ for (a) logistic regression on the Adult dataset and (b) the COMPAS algorithm using different $P_{M, \theta}(y|s, \theta)$ models, with respect to the number of data instances, averaged over 10 bootstrap data samples. The dotted blue line indicates Bayesian ensemble approach.}
		\label{fig:biasSFPerInstance} 
\end{figure*}
\begin{figure*}[t]
        \begin{minipage}{\textwidth}
            \centering Differential Fairness
        \end{minipage}
		\centerline{\includegraphics[width=0.9\textwidth]{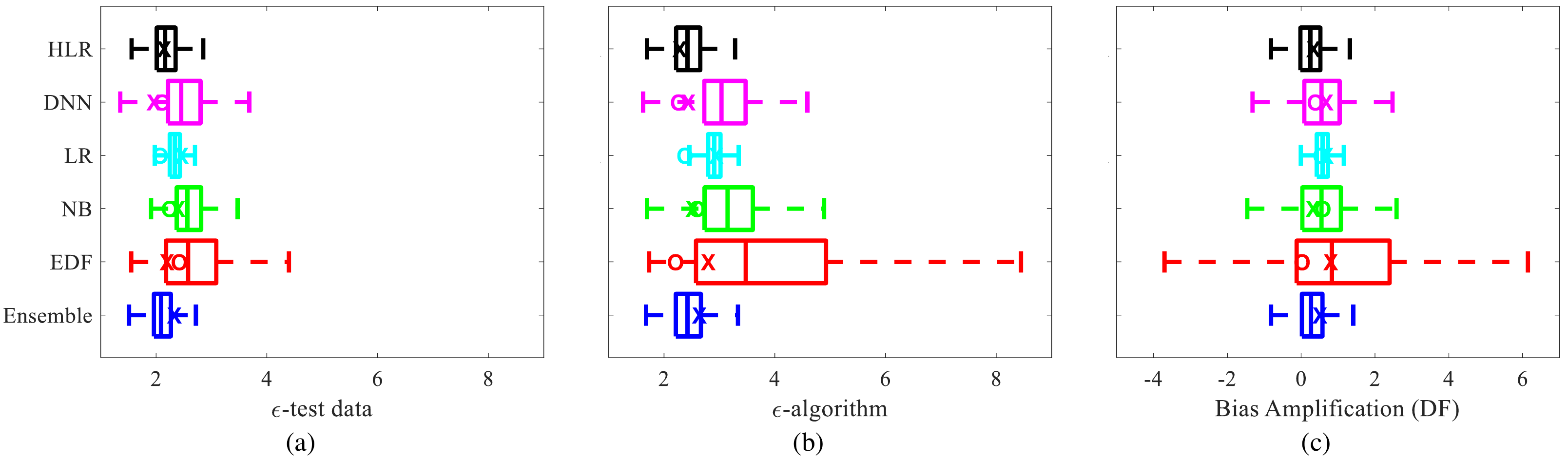}}
        \begin{minipage}{\textwidth}
            \centering Subgroup Fairness
        \end{minipage}
		\centerline{\includegraphics[width=0.9\textwidth]{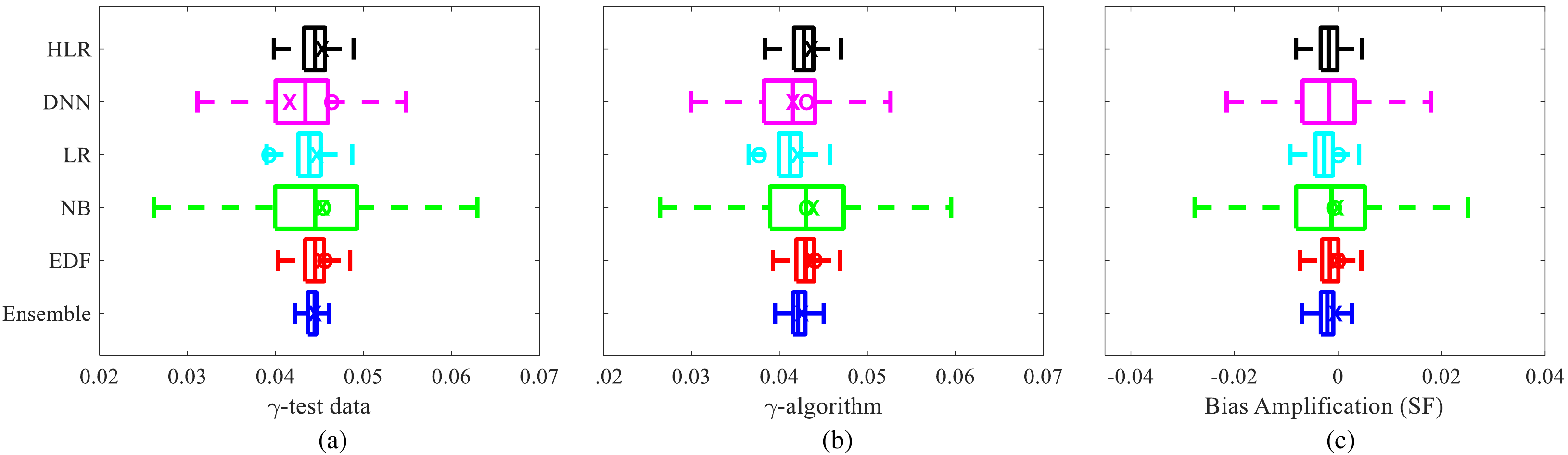}}
		\caption{\small Differential fairness (top) and Subgroup fairness (bottom) estimates using PE and variational posterior distribution of FB to model uncertainty, Adult dataset: Fairness estimates on (a) true label of data, (b) logistic regression $M(x)$-relabeled data, and (c) bias amplification by $M(x)$. The ``O'' and ``X'' on top of the box-plots indicate estimates from PE and the posterior predictive distribution of FB models, respectively.}
		\label{fig:df-sf-box-adult} 
\end{figure*}
Our proposed HLR-FB model showed consistently stable behavior in all experiments, even with a very small number of instances, producing estimates of $\epsilon$ and $\gamma$ which were similar to the final predictions of all models.  The variance in the estimates of fairness was substantial for several models, but averaging over bootstrap samples mitigated this to some degree. To illustrate this, we randomly pick a bootstrap data sample at each data instance, instead of averaging over bootstrap samples (Figure~\ref{fig:RandomepsilonAndGammaPerInstance}). In this figure, we still average over 10 bootstrap samples for the Bayesian Ensemble method, as in Figure~\ref{fig:epsilonAndGammaPerInstance}, as a reference to compare to the other models. Although all other models degrade somewhat in terms of estimation stability in this setting due to variance in their estimation, HLR-FB still enjoys relatively consistent and stable performance for both the $\epsilon$-DF and $\gamma$-SF metrics. 

In Figure~\ref{fig:biasSFPerInstance}, we show the results of measuring $(\gamma_2 - \gamma_1)$-$SF$ bias amplification, defined similarly to the DF bias amplification metric (Definition \ref{def:biasAmplification}).  We once again average over 10 bootstrap samples, varying the number of data instances for both Adult and COMPAS datasets. The results are similar to the results we obtained for $(\epsilon_2 - \epsilon_1)$-$DF$ (Figure~\ref{fig:biasDFPerInstance}). For both datasets, HLR-FB performs similarly to the Bayesian ensemble method.

Finally, we also conducted a case study with the Adult dataset where we estimated the intersectional fairness metrics, and their uncertainty via the variational posteriors (Figure~\ref{fig:df-sf-box-adult}). These results are in line with those of the COMPAS case study (Figure~\ref{fig:df-sf-box-compas}). They indicate that the direction of bias amplification is almost certainly positive for DF, however the bias amplification is roughly symmetric about $0$ for the SF metric. 

\section{Appendix: Related Work}

Bayesian modeling of fairness has been performed by \citet{simoiu2017problem} in the context of stop-and-frisk policing.  They model risk probabilities within each protected category, and require algorithms (or people, such as police officers) to threshold these probabilities at the same points when determining outcomes.

\citet{kusner2017counterfactual} use Bayesian inference on causal graphical models for fairness.  Under their \emph{counterfactual fairness} definition, changing protected attributes $A$, while holding things which are not causally dependent on $A$ constant, will not change the predicted distribution of outcomes.

As an alternative to the Bayesian methodology, adversarial methods are another strategy for managing uncertainty in a fairness context. For example, \citet{beutel2017data} apply this approach to the setting of ensuring fairness given a limited number of observations in which demographic information is available.

In a legal context, and before there was substantial research on fairness in AI, which was not their focus, \citet{roth2006modeling} and \citet{collins2008testing} 
studied various frequentist hypothesis testing methods for the 80\% rule \citep{eeoc1966guidelines} in the small data regime. These authors pointed out the dangers of determining adverse impact discrimination with small data and without proper statistical care.  Although their emphasis was not on intersectionality, AI fairness, or Bayesian methods, these papers are important precursors to our work. 

\end{document}